\documentclass[11pt,a4paper]{article}
\usepackage[hyperref]{emnlp-ijcnlp-2019}
\usepackage{times}
\usepackage{latexsym}
\usepackage{url}
\usepackage{subfigure}
\usepackage{tikz}
\usetikzlibrary{trees}
\usepackage{amsmath}
\usetikzlibrary{graphs}
\usepackage{amsfonts}
\usepackage{multirow}
\usepackage{makecell}
\usepackage{colortbl}

\aclfinalcopy 

\title{Recurrent Graph Syntax Encoder for Neural Machine Translation}

\author{%
  Liang Ding\quad\quad Dacheng Tao\\
  UBTECH Sydney AI Centre, SCS, FEIT,
  University of Sydney\\
  {\tt ldin3097@uni.sydney.edu.au, dacheng.tao@sydney.edu.au}
  }
\date{}

\begin{document}
\maketitle
\begin{abstract}
  Syntax-incorporated machine translation models have been proven successful in improving the model's reasoning and meaning preservation ability. In this paper, we propose a simple yet effective graph-structured encoder, the Recurrent Graph Syntax Encoder, dubbed \textbf{RGSE}, which enhances the ability to capture useful syntactic information. The RGSE is done over a standard encoder (recurrent or self-attention encoder), regarding recurrent network units as graph nodes and injects syntactic dependencies as edges, such that RGSE models syntactic dependencies and sequential information (\textit{i.e.}, word order) simultaneously. Our approach achieves considerable improvements over several syntax-aware NMT models in English$\Rightarrow$German and English$\Rightarrow$Czech translation tasks. And RGSE-equipped big model obtains competitive result compared with the state-of-the-art model in WMT14 En-De task. Extensive analysis further verifies that RGSE could benefit long sentence modeling, and produces better translations.
\end{abstract}

\section{Introduction}
Neural machine translation (NMT), 
proposed as a novel end-to-end paradigm~\cite{D13-1176,seq2seq,rnnsearch,convs2s,gnmt,transformer},
has obtained competitive performance compared to statistical machine translation (SMT). Although the attentional encoder-decoder model can recognize most of the structure information, there is still a certain degree of syntactic information missing, potentially resulting in syntactic errors~\cite{shi2016does,linzen2016assessing,raganato2018analysis}.
Researches on leveraging explicit linguistic information have been proven helpful in obtaining better sentence modeling results~\cite{kuncoro2018lstms,strubell2018linguistically}. We therefore argue that explicit syntactic information (here we mainly focusing on utilizing syntactic dependencies) could enhance the translation quality of recent state-of-the-art NMT models.

Existing works incorporating explicit syntactic information in NMT models has been an active topic~\cite{stahlberg-etal-2016-syntactically,aharoni-goldberg-2017-towards,li-EtAl:2017:Long,D17-1209,chen-EtAl:2017:Long6,wu2017improved,Wu:2018:DNM:3281228.3281242,zhang2019syntax}. However, they are mostly sophisticated in designing and have not proven their effectiveness in the latest structure(\textit{i.e.,} Transformer). Recent studies have shown that graph neural networks (GNN)~\cite{scarselli2009graph} and its variants (\textit{e.g.,} graph convolutional network(GCN)~\cite{kipf2016semi}, graph recurrent network(GRN)~\cite{zhang2018sentence}) have benefited natural language representation~\cite{battaglia2016interaction,hamilton2017inductive,marcheggiani-titov:2017:EMNLP2017,N18-2078,P18-1026,song2018graph,song2018n,song2019semantic} with high interpretability for non-Euclidean data structures. Despite these apparent successes, it still suffers from a major weakness: their graph layer assumes that nodes are distributed independently without explicit word order (\textit{i.e.,} nodes within a graph layer essentially acting as non-recursive quasi-RNNCell in formula), overlooking internal sequential knowledge.

To overcome above issues, we presents a novel Recurrent Graph Syntax Encoder (RGSE), casting nodes in graph layer as RNNCells, which has the central approach of capturing syntactic dependencies and word order information simultaneously. Specifically, RGSE first receives each word's representation of the original encoder and then makes each RNN node in RGSE layer obtain its dependency nodes (\textit{i.e.,} words dependencies from the original encoder) and the previously hidden state in the RGSE layer. RGSE could not only flexibly deployed over original encoder of recurrent NMT but also can be utilized on Transformer. Furthermore, RGSE could enhance the inductive learning ability for models since more syntax connections are provided to guide source-side meaning preservation and target-side word prediction. Our main contributions are summarized as follows:

\begin{itemize}
\item We propose a simple yet effective representation method, RGSE for NMT, which is done over a standard encoder (recurrent or transformer) and could informs the NMT model with comprehensive syntactic dependencies. The edge-wise integration, on the other hand, enables attentional decoder to pick essential source words for prediction.

\item We develop a novel Transformer architecture that alternates the self-attention component with RGSE in the lower layers. The alternation allows the encoder capture more prior knowledge (\textit{i.e.,} syntactic dependency information), improving the representation and induction ability for Transformer. The gated residual connection, on the other hand, yields fast convergence speed.

\item Experiments on English-German (standard WMT14 and WMT16 News Commentary V11) and English-Czech (WMT16 News Commentary V11) translation tasks show consistent improvements over several strong syntax aware baselines, validating the effectiveness and universality of RGSE.

\end{itemize}
We conduct extensive experiments with different setups to find the optimal setting: having RGSE in one direction and in both directions; integrating incoming edges with different functions; and including dependencies from past (previous) or future (following) words. For the Transformer-based NMT~\cite{transformer}, empirical experiments on validation set showed that replacing the self-attention component in the lower layers with RGSE performed better, probably because Transformer tend to capture some complex and long dependencies at higher layers but showing relatively poor dependencies modeling ability in lower layers~\cite{raganato2018analysis}. In doing so, our bidirectional edge-wise RGSE-equipped NMT models could achieve further improvements over several strong syntax-aware NMT models.

\section{Background}
\label{BG}

Our model is based on the sequence-to-sequence framework~\cite{seq2seq,rnnsearch,luong2015multitask,transformer}. In NMT, we normally employ an encoder with the assumption that it can adequately represent the source sentence. Then, the decoder can autoregressively predict each target word. This section will briefly review the ``\textit{neural encoder}'' and ``\textit{graph syntax encoder}'' respectively.

\subsection{Neural Encoder}
The NMT encoder intents to summarize the source semantics and dependencies such that the decoder could generate them with target words. We will describe two kinds of popular encoders (\textit{i.e.,} \textit{RNN encoder}~\cite{D13-1176,seq2seq,rnnsearch,gnmt} and \textit{Self-Attentional Transformer encoder}~\cite{transformer}) in succession.

\begin{figure}[htb]
    \centering
    \includegraphics[width=.49\textwidth]{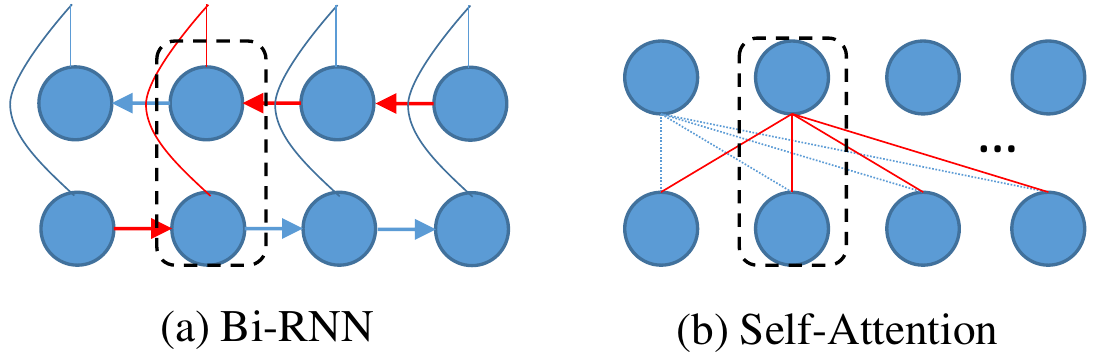}
    \caption{Simplified illustration of BiRNN (a) and Self-Attention encoder (b), the black dotted rectangle represents the node of current step, and the red line denotes the hidden state that can be perceived at that moment.}
    \label{fig:encoder}
\end{figure}

\subsubsection{RNN encoder}
As is shown in Fig.~\ref{fig:encoder}, given an RNN encoder,  we can bidirectionally model a sentence as follows:
\begin{equation*}
\begin{split}
    \overrightarrow{h_t} = 
    \overrightarrow{RNN}({\rm E}_{src}{\rm x}_t,\overrightarrow{h_{t-1}}) \textrm{,}\\
    \overleftarrow{h_t} =
    \overleftarrow{RNN}({\rm E}_{src}{\rm x}_t,\overleftarrow{h_{t+1}})
\end{split}
\end{equation*}
where ${\rm x}_t \in \{0,1\}^{|V_{src}|}$ is the one-hot vector and ${\rm E}_{src}{\rm x}_t \in \mathbb{R}^{|d_{emb}|}$ indicates the embedding of the $t_{th}$ source word. Then, the above two vectors will be concatenated as $\widetilde{h_t}=[\overrightarrow{h_t};\overleftarrow{h_t}]$ to represent the contextual information.

\subsubsection{Self-Attentional Transformer encoder}
On the other hand, self-attention in Transformer~\cite{transformer} allows the encoder to model sentence representation parallelly. Each sub-layer consists of two main components: self-attention layer and feed-forward network. the self-attention layer receives a list of vectors as inputs(See Fig.~\ref{fig:encoder}). For any sequence with length $L$ containing $m$ steps, the $t_{th}$ word in the $i_{th}$ layer can be denoted as:
\[
    z_t^{(i)}=\begin{cases}\sum\limits_{m=1}^{L}softmax(\frac{<q_t,k_m>}{\sqrt{d}})v_t&i\geq1 \\&\\\sqrt{d}\times{\rm E}_{src}{\rm x}_t+PE_t & i=0\end{cases}
\]
where $q_t$, $k_m$ and $v_t$ are equal to $\theta_{q}h_t^{(i-1)}$, $\theta_{k}h_m^{(i-1)}$, and $\theta_{v}h_t^{(i-1)}$; specifically, they refer to the query, key, and value in the $i-1$ layer and $\theta$ stands for trainable weight matrix. The similarity between $query$ and $key$ can be evaluated by dot-production attention. $PE$ is the fixed position embedding and its dimension $d$ is consistent with the word embedding, defined as:
\begin{equation*}
\begin{aligned}
    PE_{t,j}&=\sin(t/10000^{2j/d}) \\
    PE_{t,2j+1}&=\cos(t/10000^{2j/d})
\end{aligned}
\end{equation*}

\subsection{Graph Syntax Encoder}
  In NLP tasks, besides consecutive word order information, non-local neighbor relations (\textit{e.g.,} dependency relations) are also crucial. To inform the model with non-local information, the graph structure networks are employed~\cite{marcheggiani-titov:2017:EMNLP2017,D17-1209,song2018graph,P18-1026,song2019semantic}.
  
  The syntax GCN layer is adopted to connect words with semantic dependencies over the original encoder~\cite{D17-1209}. Formally, the hidden state of node $\nu$ with neighbor dependency nodes collection $\mu \in N(\nu)$ can be described as:
  \begin{equation*}
      h_{\nu}=\rho\left(\sum_{\mu}W_{dir(\mu,\nu)}h(\mu)+b_{lab(\mu,\nu)} \right)
  \end{equation*}
  where $dir$ and $lab$ refer to directionality and labels, $\rho$ is the non-linear activation function, and trainable parameters were defined as $W$ and $b$. 
  
  To capture non-local dependencies while propagating information in the recurrent network, we can sum the incoming and outgoing edges as input. For example, ~\newcite{song2018graph,song2019semantic} denotes the inputs for the input gate and output gates as $x_{\nu}^i = \sum_{\mu \in N_{in}({\nu})} x_{\mu}$, $x_{\nu}^o = \sum_{\mu \in N_{out}({\nu})} x_{\mu}$ in the transitioning process of the graph-state LSTM with a similar operation for hidden states $h_{\nu}^i$ and $h_{\nu}^o$, where incoming and outgoing neighbors of $\nu$ are denoted by $N_{in}(\nu)$ and $N_{out}(\nu)$. 
  
Similarly,~\newcite{P18-1026} applied gated GNN layer that directly received word and positional embeddings to represent dependency information.

With these effective graph structure strategies, the encoder can incorporate explicit non-local dependency information. However, their graph layers are essentially different with ours, assuming that nodes in each graph layer are orthogonal (\textit{i.e.,} there is no sequential information propagation between nodes). Our RGSE could model both word order information and syntactic dependencies.

\section{Recurrent Graph Syntax Encoder}
\label{RGSE}

Here, we present RGSE and explain how it is assembled in Recurrent NMT and Transformer.

\subsection{RGSE for Recurrent NMT}
\label{ssec:RGSE4RNMT}

As mentioned above, we choose RNN as the activation cell (more specifically GRU, as we follows the Recurrent NMT model settings of~\newcite{rnnsearch}), which means RGSE layer not only conveys non-local dependencies but also records the consecutiveness. Meanwhile, inspired by~\newcite{wu2017improved,Wu:2018:DNM:3281228.3281242} where they bidirectionally model the in-order sequence of the dependency tree, we believe that bidirectional propagation will enhance its representation ability.

For any input graph $G=\langle V,E \rangle$, we define vectors $\widetilde h$, $\overrightarrow s$ and $\overleftarrow s$ for each word $\nu \in V$, where $\widetilde h$ is from the previous encoder, bidirectional $s$ represent forward and backward states in RGSE layer, and $|V|$ is the length of the source sentence. For any pair of dependent words $w_i\mapsto w_j$ in a sentence, nodes $\overrightarrow{s_j},\overleftarrow{s_j}$ will be activated; concurrently, two edges $\xi_{(\widetilde h_i, \overrightarrow s_j)}$ and $\xi_{(\widetilde h_i, \overleftarrow s_j)}$ will be generated. All incoming edges $\xi \in{E_{in}(s_j)}$ for node $s_j$ are integrated through three types of functions:
\[\phi(s_j)=\begin{cases}
     \sum\limits_{E_{in}(s_j)}h_{\xi_{(\widetilde h_i, s_j)}} &sum\\ \sum\limits_{E_{in}(s_j)}1/{|N_{in}(s_j)|}\cdot h_{\xi_{(\widetilde h_i, s_j)}} &average\\ 
     \sum\limits_{E_{in}(s_j)} W_{h_{\xi_{(\widetilde h_i, s_j)}}} \cdot h_{\xi_{(\widetilde h_i, s_j)}} &gated
     \end{cases}
\]
          
          
where hidden vector $h$ is the value of $\widetilde h_i$ and $W$ is trainable gating parameter. Then, the propagation processes in bidirectional RGSE are:
\begin{equation*}
    \begin{split}
        \overrightarrow z_t &= \sigma(\overrightarrow W_z \overrightarrow s_{t-1} + \overrightarrow U_z \phi(\nu_t) + \overrightarrow b_z) \textrm{,} \\
        \overleftarrow z_t &= \sigma(\overleftarrow W_z \overleftarrow s_{t+1} + \overleftarrow U_z \phi(\nu_t) + \overleftarrow b_z) \textrm{,} \\
        \overrightarrow r_t &= \sigma(\overrightarrow W_r \overrightarrow s_{t-1} + \overrightarrow U_r \phi(\nu_t) + \overrightarrow b_r) \textrm{,} \\
        \overleftarrow r_t &= \sigma(\overleftarrow W_r \overleftarrow s_{t+1} + \overleftarrow U_r \phi(\nu_t) + \overleftarrow b_r) \textrm{,} \\
        \overrightarrow s_t^{\prime} &= \tanh(\overrightarrow W_h \phi(\nu_t) + \overrightarrow U_h(\overrightarrow r_t \overrightarrow s_{t-1})) \textrm{,} \\
        \overleftarrow s_t^{\prime} &= \tanh(\overleftarrow W_h \phi(\nu_t) + \overleftarrow U_h(\overleftarrow r_t \overleftarrow s_{t+1})) \textrm{,} \\
        \overrightarrow s_t &= \overrightarrow z_t \cdot \overrightarrow s_{t-1} + (1-\overrightarrow z_t)\cdot \overrightarrow s_t^{\prime} \textrm{,} \\
        \overleftarrow s_t &= \overleftarrow z_t \cdot \overleftarrow s_{t+1} + (1-\overleftarrow z_t)\cdot \overleftarrow s_t^{\prime} \textrm{,} \\
        \eta_t &= \tau(\overrightarrow s_t, \overleftarrow s_t, \widetilde h_t) \\
    \end{split}
\end{equation*}
where $\overrightarrow s_t$ and $\overleftarrow s_t$ are the outputs of forward and backward RGSE, $\eta_t$ is the final state of encoder at time $t$, and $\tau$ refers to the residual concatenation. Here we employ normal residual connection $\tau_n(\cdot)$~\cite{he2016deep}:
\begin{equation*}
    \tau_n(\overrightarrow s,\overleftarrow s, \widetilde h) = \text{concat}(\overrightarrow s + \widetilde h,\overleftarrow s + \widetilde h)
\end{equation*}
and another alternative gated residual connection: 
\begin{equation*}
    \begin{split}
        \tau_g(\overrightarrow s,\overleftarrow s, \widetilde h) = \text{concat}(&\lambda_1 \overrightarrow s + (1-\lambda_1)\widetilde h,\\& \lambda_2 \overleftarrow s + (1-\lambda_2)\widetilde h)
    \end{split}
\end{equation*}
where $\lambda$ can be calculated as:
\begin{equation*}
    \lambda = \sigma(\omega\cdot s + \psi\cdot h)\label{lambda calculation}
\end{equation*}
$\omega$ and $\psi$ are gating parameters. To further distinguish which directional RGSE layer is better and whether using past information or future information alone could enhance RGSE performance, we design the following four RGSE models: (In the following models, we give an example sentence ``\textit{monkey likes eating bananas}'', in which there exists three pairs of dependencies: ``$monkey \mapsto like$, $eating \mapsto like$, and $bananas \mapsto eating$'')\\
\begin{figure}
    \centering
    \scalebox{1}{\includegraphics[width=6cm]{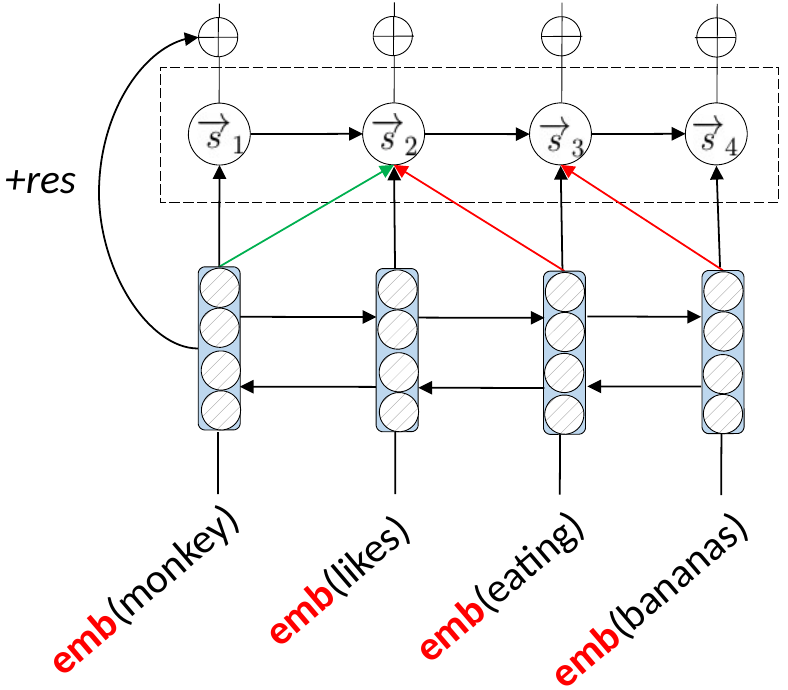}}
    \caption{Uni-layer RGSE upon RNMT encoder.}
    \label{fig:RGSE_uni-layer}
\end{figure}
\textbf{(\romannumeral1) Forward RGSE}: 
Fig.~\ref{fig:RGSE_uni-layer} illustrates the uni-layer forward-RGSE model (\verb|forward-RGSE|, within the dashed rectangle). The original encoder reads embedded word vectors before connecting with the RGSE layer. As mentioned above, the integration function $\phi(\cdot)$ was employed for each node to properly capture the incoming edges, where green edges represent past information and red edges show future information. For example, node $\overrightarrow s_2$ in Fig.~\ref{fig:RGSE_uni-layer} receives both RGSE hidden state $\overrightarrow s_1$ and dependency information, which includes current word ``\textit{like}'' from the original encoder, past information ``\textit{monkey}'', and future information ``\textit {eating}''.  
After original encoding and RGSE modeling, uppermost layer $\tau(\cdot)$ will combine them position-wisely.

\begin{figure}
    \centering
    \includegraphics[width=6.6cm]{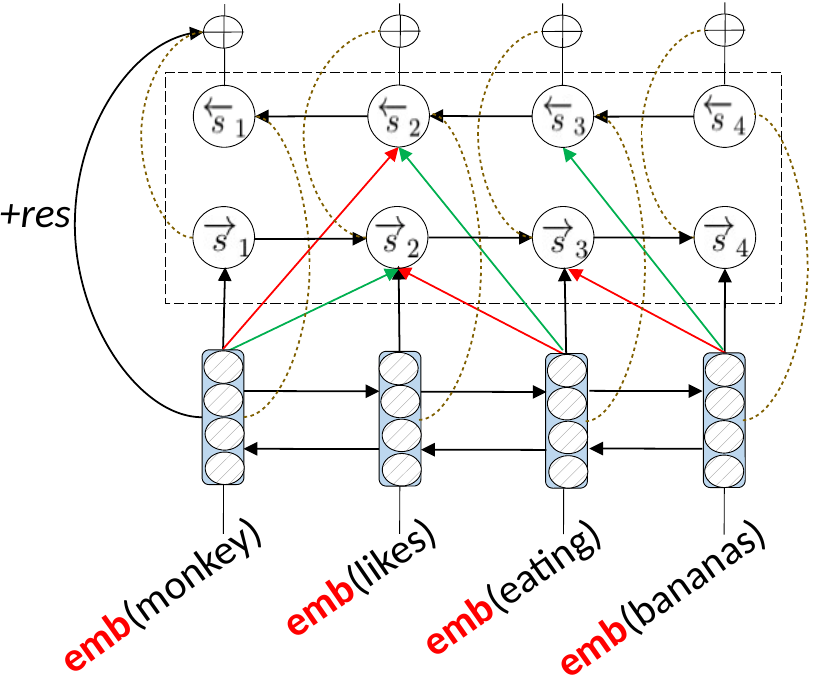}
    \caption{Bidirectional RGSE upon RNMT encoder.}
    \label{fig:RGSE_bi-layer}
\end{figure}
\textbf{(\romannumeral2) Bidirectional Total RGSE}:
In order to make full use of the property of recurrent network, we intuitively add a reverse RGSE layer. Fig.~\ref{fig:RGSE_bi-layer} shows the bidirectional RGSE(called \verb|bi-total-RGSE|). Both forward and backward RGSE layers read the hidden state and dependencies from the original encoder.

\begin{figure*}[htb]
    \centering
     \scalebox{0.89}{\includegraphics[width=0.96\textwidth]{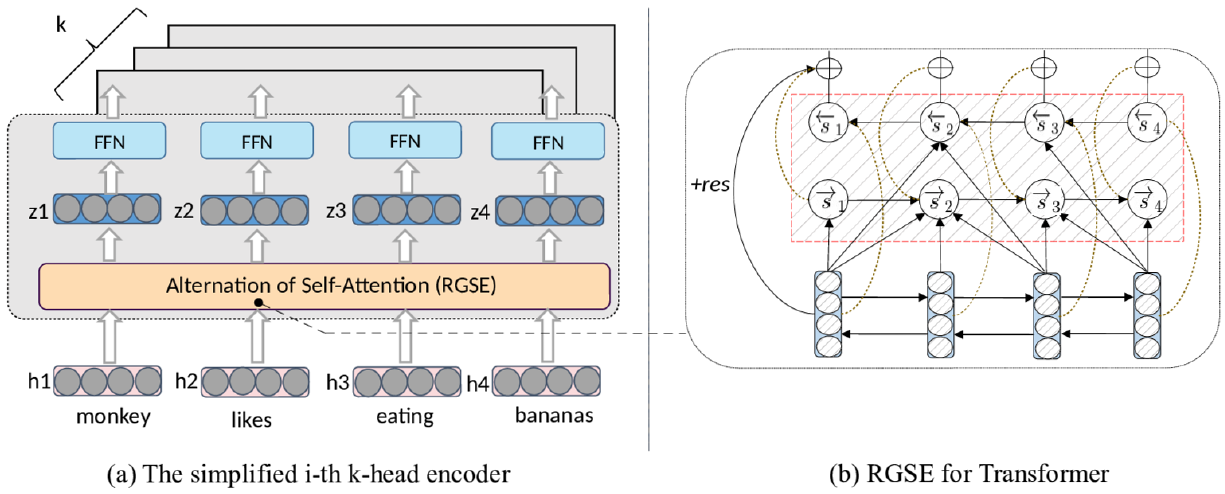}}
    \caption{Illustration of RGSE-based Transformer encoder. (a) is the simplified $i_{th}$ encoder layer of the Transformer, which receives the hidden state of each word from previous layer. ${\rm FFN}$ refers to the feed forward networks and $k$ is multi-head number. (b) shows how the bi-total-RGSE upon bi-GRU layer replace the self-attention.}
    \label{fig:self-attention encoder}
\end{figure*}

\textbf{(\romannumeral3) Bidirectional Past RGSE}: Bidirectional past RGSE (\verb|bi-past-RGSE|), unlike \verb|bi-total-RGSE|, only gathers past edges (marked as green arrows). For example, although the backward node $\overleftarrow s_2$ has dependent relationships with ``\textit{monkey}'' and ``\textit {eating}'', $\overleftarrow s_2$ only reads dependent edge $\xi_{(h_3,\overleftarrow s_2)}$ because, in reverse order, ``\textit{eating}'' is the past information of ``\textit{like}''.

\textbf{(\romannumeral4) Bidirectional Future RGSE}:
Contrary to \verb|bi-past-RGSE|, \verb|bi-future-RGSE| reads future dependencies (marked as red arrows) only.

\subsection{RGSE for Transformer}
\label{ssec:RGSE4Transformer}

Although Transformer has achieved the state-of-the-art performance, it possesses the innate disadvantage on sequential modeling. Taking sentence ``\textit{I bought a new \textbf{book} with a new friend}'' for instance, during modeling word ``\textbf{book}'' with positional embedding removed self-attention, it will pay same attention on two ``\textbf{new}'' while the true case is it only needs to pay attention to the first ``\textbf{new}''. To improve this issue, \newcite{shaw2018self} proposed relative position for Transformer and \newcite{yang2018modeling} introduced Gaussian bias into encoder layers as prior constraint. 

It is linguistic intuition that the syntax information could enhance the representation ability. \newcite{P18-1167} reported that replacing the self-attention layer with RNN in the encoder could deliver comparable results to the vanilla Transformer. We assume that adding \verb|bi-total-RGSE| to the bi-GRU-replaced Transformer (see Fig.~\ref{fig:self-attention encoder}) could be helpful. In addition, we also investigated which level of layers benefit most from RGSE in experiments.

\section{Experiments}
\label{EXP}

The aims of experiments are (1) finding the optimal structure of RGSE on validation data set (2) proving the superiority of RGSE over existing tree\&graph-structure syntax-aware models (3) assessing the effectiveness of RGSE-based Transformer compared with several SOTA models.

\subsection{Setup}
To compare with the results reported by previous works~\cite{D17-1209,P18-1026} under the recurrent NMT scenario, we conduct experiments on News Commentary V11 corpora from WMT16\footnote{\url{http://www.statmt.org/wmt16/translation-task.html}}, comprising approximate 226K En-De and 118K En-Cs sentence pairs respectively, where the data and settings are consistent with them. We employ SyntaxNet\footnote{\url{https://github.com/tensorflow/models/tree/master/research/syntaxnet}} to tokenize and parse English side data while German and Czech corpora are segmented by byte-pair encodings (BPE)~\cite{sennrich2016neural}, where we use 8K BPE merges to avoid OOV problem. Further preprossing details follow~\newcite{D17-1209}. 
As comparison, we reimplemented SE-NMT~\cite{wu2017improved,Wu:2018:DNM:3281228.3281242} where they employed MLP function to concatenate four hidden states (forward/reverse in-order traversal, pre-order traversal and post-order traversal from syntax dependency tree) and trained Tree2Seq~\cite{chen-EtAl:2017:Long6} model with their released code\footnote{\url{https://github.com/howardchenhd/Syntax-awared-NMT}}. We also conduct Transformer-based experiments on NC-v11 dataset as reference. 

To assess the effectiveness of RGSE on advanced Transformer-based model~\cite{transformer} and fairly compare with other state-of-the-art models~\cite{shaw2018self,yang2018modeling}, we implement RGSE equipped Transformer on top of an open-source toolkit OpenNMT\footnote{\url{https://github.com/OpenNMT/OpenNMT-py}}~\cite{opennmt}. We followed~\newcite{transformer} to set the configurations and report results on standard WMT14 English$\Rightarrow$German task\footnote{\url{https://nlp.stanford.edu/projects/nmt}}, which consists of 4.5M sentences pairs. Here we processed the BPE with 32K merge operations for both language pairs\footnote{The label will remain on each substring if a word is splitted by BPE.}. For fair comparison, here we also implement the key idea of~\newcite{D17-1209} into Transformer framework in two ways: one is similar to our approach, replacing self-attention with ``BiRNN+GCN'', another is simply adding GCN upon self-attention layer (followed the same edge dropout rate 0.2) before feed-forward-network processing. All models were trained on 6 \verb|NVIDIA V100| GPUs, where the batch size is 4096 tokens. Note thea the 4-gram NIST BLEU score~\cite{Papineni:2002:BMA:1073083.1073135} is applied as the evaluation metric for all models. 

\subsection{Ablation Study}
To achieve the aim (1), we first evaluated the effects of internal functions on RGSE, then assessed which level of layers is most beneficial to be applied RGSE on Transformer. The results are reported on validation set and trained with News Commentary V11 English$\Rightarrow$German corpora.

\begin{figure}[htb]
\centering
    \includegraphics[width=6.9cm,height=5.6cm]{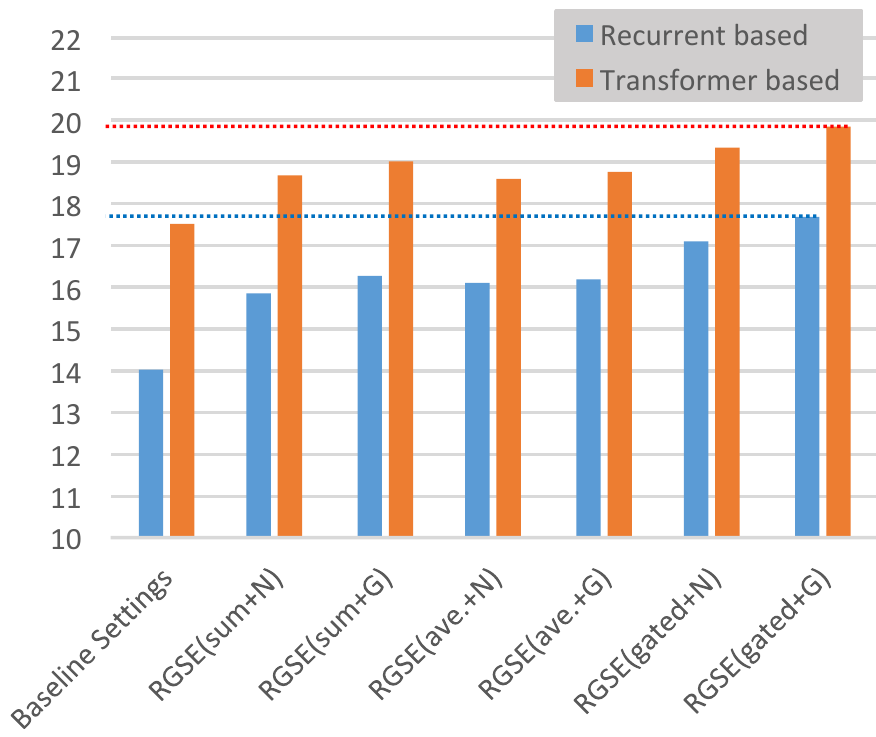}
    \caption{Validation BLEU of different settings for En-De NC-v11, where the baselines are Recurrent and Transformer NMT. Beyond baselines, other RGSE systems mean bi-total-RGSE structure. \textit{sum}, \textit{ave.}, and \textit{gated} refer to three types of integration function $\phi(\cdot)$, symbols +$\mathcal{N}$ and +$\mathcal{G}$ signify normal residual connection $\tau_n(\cdot)$ and gated residual connection $\tau_g(\cdot)$, respectively. Note that RGSE component applied to every layer for Transformer-based model in ablation study.}
    \label{fig:function_comparison}
\end{figure}

\paragraph{Effects of internal functions.} 
Which integration function $\phi(\cdot)$ is more helpful? What kind of residual connection method $\tau(\cdot)$ helps to improve translation? Fig.~\ref{fig:function_comparison} illustrates the performance of different integration functions on the NC-v11 En-De validation set. Comparisons show that the combination of the edge-wise integration function and gated residual connection most benefits our task. Therefore, the following experiments will use the edge-wise integration function and gated residual connection as the default configuration.

\paragraph{Effects of different level of layers.} ~\newcite{anastasopoulos2018tied} stated that high-level layers exploit more structure information and more long-distance dependencies than lower layers. We thus design ablation study to investigate if it is necessary to deploy RGSE on every layer.  As is shown in Tab.~\ref{tab:layers}, modeling the first three layers with RGSE in Transformer can achieve the best performance. This result is consistent with~\newcite{yang2018modeling}'s findings, validating our assumption.

\begin{table}[t]
\centering
    \begin{tabular}{c|c||c|cc}
    \textbf{\#} &  \textbf{Layers} & \textbf{Speed} & \textbf{Val.} & $\Delta$\\
    \hline\hline
    1&[1-6]&1.52&19.80&-\\
    \hline
    2&[1-1]&1.41&19.89&+0.09\\
    3&[1-2]&1.43&19.94&+0.14\\
    4&[1-3]&1.43&\textbf{20.01}&\textbf{+0.21}\\
    5&[1-4]&1.46&19.89&+0.19\\
    \hline
    6&[4-6]&1.42&19.91&+0.11\\
    \end{tabular}
    \caption{Different settings that employed RGSE on different layer combinations in Transformer. ``speed'' denotes training speed measured in steps per second.}
    \label{tab:layers}
\end{table}

\subsection{Main Results}
To achieve aim (2), we first report and analyze the BLEU scores on NC-v11 En-De and En-Cs test sets. Then we would compare with several SOTA systems on standard WMT14 En-De dataset to accomplish aim (3).

\begin{table*}[t]
    \centering
    \begin{tabular}{l|l|ll|ll}
    \multirowcell{2}{}&\multirowcell{2}{\textbf{System}}&\multicolumn{2}{c|}{En-De}&\multicolumn{2}{c}{En-Cs}\\
    \cline{3-6}
    &&BLEU&\#para.&BLEU&\#para.\\
    \hline\hline
    \multirowcell{5}{\textit{existing works}} & PB-SMT~\cite{P18-1026}&12.8&n/a&8.6&n/a\\
    &Bi-RNN~\cite{D17-1209}&14.9&n/a&8.9&n/a\\
    &Bi-RNN + GCN~\cite{D17-1209}&16.1&n/a&9.6&n/a\\
    &Tree2Seq~\cite{chen-EtAl:2017:Long6}&15.9&40.8M&9.4&38.1M\\
    &SE-NMT~\cite{Wu:2018:DNM:3281228.3281242}&16.4&42.5M&9.7&39.1M\\
    &Gated-GNN2S~\cite{P18-1026}&16.7&41.2M&9.8&38.8M\\
    \hline
    \hline
    \multirowcell{7}{\textit{this work}} & Bi-RNN (2 layers encoder) &15.5&62.3M&9.3&58.2M\\
    &Bi-RNN~{\rm + forward RGSE}&16.0&41.4M&9.7&39.2M\\
    &Bi-RNN~{\rm + bi past RGSE}&16.5$^\uparrow$&45.6M&10.1$^\uparrow$&42.1M\\
    &Bi-RNN~{\rm + bi future RGSE}&16.8$^\uparrow$&45.4M&10.3$^\uparrow$&42.5M\\
    &Bi-RNN~{\rm + bi total RGSE}&17.7$^\Uparrow$&52.2M&11.1$^\Uparrow$&49.8M\\
    \cline{2-6}
    &Transformer-base&18.9&80.7M&11.6&76.0M\\
    &~~~{\rm +bi total RGSE}&19.8$^\Uparrow$&83.2M&12.4$^\Uparrow$&78.4M\\
    \end{tabular}
    \caption{Experiments on NC-v11 dataset. ``$\uparrow$ / $\Uparrow$'': significantly outperform their counterpart ($p < 0.05/0.01$).}
    \label{tab:result_ncv}
\end{table*}

Tab.\ref{tab:result_ncv} proves the effectiveness of RGSE model on NC-v11 dataset and its superiority over existing works (both tree-based and graph-based syntax-aware models). Unsurprisingly, SMT performs the worst. The tree-based models (\textit{i.e.,}Tree2Seq\cite{chen-EtAl:2017:Long6}, SE-NMT\cite{Wu:2018:DNM:3281228.3281242}) and graph-based models (\textit{i.e.,}BiRNN+GCN\cite{D17-1209}, Gated-GNN\cite{P18-1026}) easily outperform the SMT and BiRNN as expected. In this work, for proving that the performance gains are not due to increased number of parameters, we employ Bi-RNN with 2 layers encoder as baseline system as its parameter scale is larger than ours, and besides this, other settings all employed 1 layer BiRNN encoder. Results of several GRSE models confirm that forward-RGSE, bi-past-RGSE, bi-future-RGSE and bi-total-RGSE are progressively improve translation. Most notably, it outperforms the strong baseline by +2.2 and + 1.8 points on En-De and En-Cs tasks respectively and bi-total-RGSE significantly exceeds existing syntax-aware models. To verify universality of RGSE, we also conduct experiments on Transformer and compare with RGSE-equipped Transformer. Experiments show adding RGSE could make the Transformer obtain +0.9 BLEU on En-De and +0.8 BLEU on En-Cs. 

In addition, we also conducted experiments on WMT14 En-De dataset to assess our model performance compared with several state-of-the-art systems. Tab.~\ref{tab:result_sota} presents recent popular models~\cite{gnmt,convs2s,transformer,shaw2018self,yang2018modeling,ahmed2018weighted,wu2019pay}. For fairly comparing with existing syntax incorporating method~\cite{D17-1209}, we reproduce BiRNN Transformer, BiRNN+GCN Transformer in lower layers ([\verb|1-3|]). Notably, RGSE-based Transformer-big model surpasses several existing powerful models, and even achieves the competitive result compared to the most advanced \textsc{DynamConv} model~\cite{wu2019pay}.

\subsection{Analysis}
We further analyze two questions in this section: (1) which type of dependency information is more important? past or future? and (2) can RGSE improve the translation quality of long sentences?
\begin{table}[t]
    \centering
    \begin{tabular}{p{4.7cm}|ll}
    \textbf{System}&BLEU&\#para.\\
    \hline
    \hline
    GNMT\cite{gnmt}&26.30&n/a\\
    ConvS2S\cite{convs2s}&26.36&n/a\\
    \hline
    Transformer-base&27.64&88.0M\\
    +Rel\_Pos\cite{shaw2018self}&27.94&88.1M\\
    +Localness\cite{yang2018modeling}&28.11&88.8M\\
    Weighted\cite{ahmed2018weighted}&28.40&n/a\\
    \hline
    Transformer-big&28.58&264.1M\\
    +Localness\cite{yang2018modeling}&28.89&267.4M\\
    Weighted\cite{ahmed2018weighted}&28.90&n/a\\
    LightConv\cite{wu2019pay}&28.90&n/a\\
    DynamConv\cite{wu2019pay}&29.70&n/a\\
    \hline
    \multicolumn{3}{c}{\textit{this work (below)}}\\
    \hline
    Transformer-base&27.65&90.2M\\
    ~~~+ GCN &27.87&90.4M\\
    ~~~+ BiRNN+GCN &27.92&91.2M\\
    ~~~+ RGSE (ours)&28.62$^\Uparrow$&91.76M\\
    \hline
    Transformer-big&28.60&272.1M\\
    ~~~+ RGSE(ours)&29.47$^\Uparrow$&278.3M\\
    \end{tabular}
    \caption{Comparing with several SOTA models on WMT14 En-De test sets. ``$\uparrow$ / $\Uparrow$'': significantly outperform their counterpart ($p < 0.05/0.01$).}
    \label{tab:result_sota}
\end{table}
\subsubsection{Past vs. Future}
Interesting results in Tab.\ref{tab:result_ncv} show that future information is somewhat more instructive compared to past information. We assume the reason is that for Subject-Verb-Object languages (\textit{e.g.,} English) future dependencies make the encoder preserve more meaningful presentations, the decoder hence can have more far-sighted predictions.

\subsubsection{Long Sentence Translation}
\begin{figure}[t]
    \centering
    \includegraphics[width=7cm,height=5.5cm]{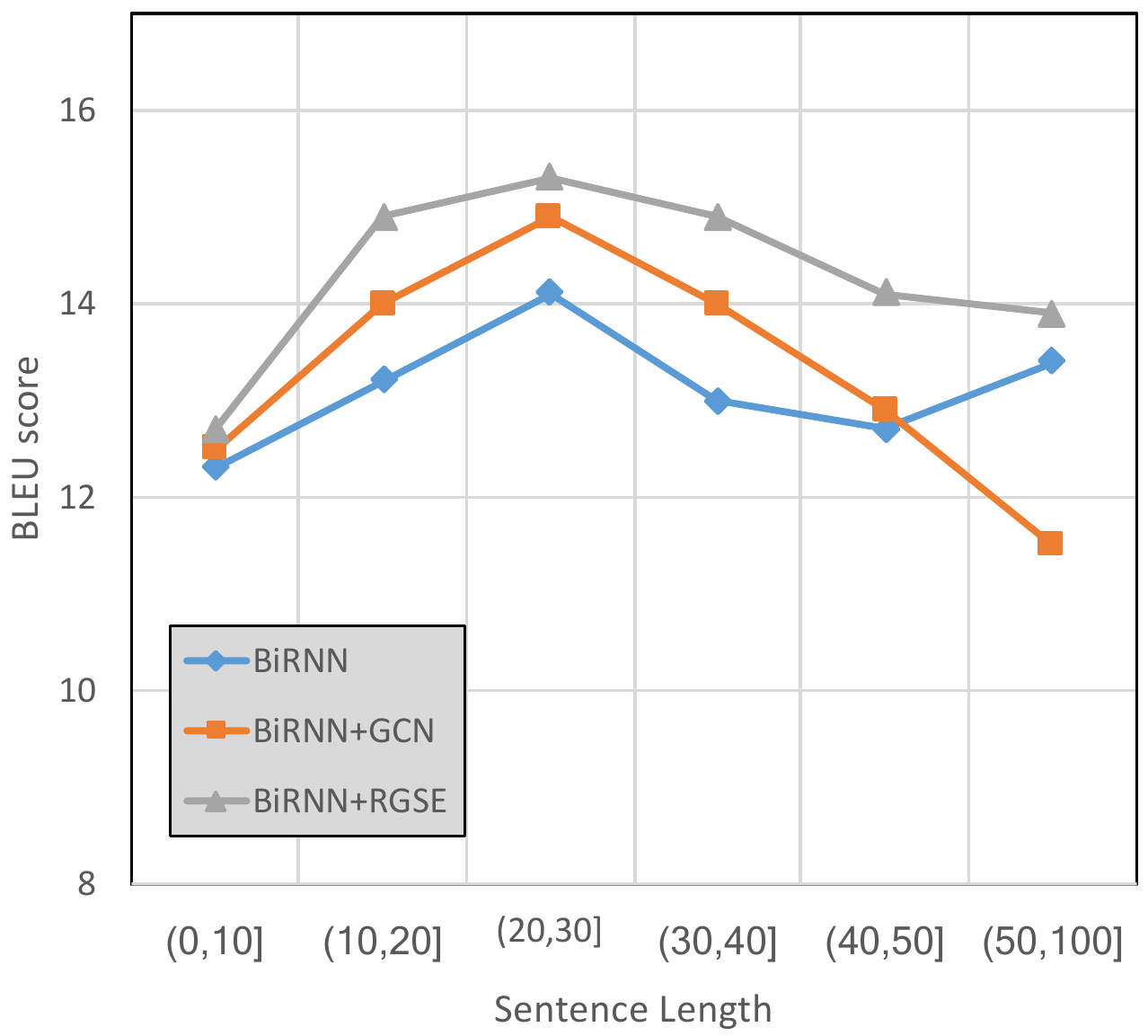}
    \caption{BLEU scores of the generated translations on NC-v11 En-De test set.}
    \label{fig:line-chart}
\end{figure}
Following~\newcite{rnnsearch}, we divide test sentences w.r.t their lengths. Fig.~\ref{fig:line-chart} indicates that RGSE-based system outperforms others when tackling long sentences, verifying our assumption that the graph nodes with recurrent dependencies can better represent long distance information.

The reason why they perform poorly when the length exceeds 50 is that the length limitation of the training corpus is 50, making it hard for the model to cope with long sentences.

\section{Related Work}

The RGSE is inspired by two research themes: 
\paragraph{Incorporating linguistic features :} Several approaches have incorporated linguistic features into NMT models since~\newcite{Tai2015ImprovedSR} demonstrated that incorporating structured semantic information could enhance the representations.~\newcite{sennrich-haddow:2016:WMT} fed the encoder cell combined embeddings of linguistic features including lemmas, subword tags, etc.~\newcite{eriguchi-hashimoto-tsuruoka:2016:P16-1} employed the tree-based encoder to model syntactic structure. ~\newcite{li-EtAl:2017:Long} showed that stitching the word and linearization of parse tree is a effective method to incorporate syntax. ~\newcite{C18-1120,P18-1116} utilized a forest-to-sequence model, which encoded a collection of packed parse trees to compensate for the parser errors, which was superior to the tree-based model. But their works does not utilize graph network to model structured data. Jointly learning of both semantic information and attentional translation is another prevalent approach that appropriately introduces linguistic knowledge. To the best of our knowledge,~\newcite{luong2015multitask} first proposed adding source syntax into NMT with a sharing encoder.~\newcite{Niehues2017ExploitingLR} trained the machine translation system with POS and named-entities(NE) tasks at the same time, gaining considerable improvements in multiple tasks.~\newcite{zhang2019syntax} concatenated the original NMT word representation and the syntax-aware word representation derived from the well-trained dependency parser. However, they considered more implicit information, overlooking the importance of explicit prior knowledge, and have not proven their effectiveness in the Transformer.

\paragraph{NMT with graph representation :}
This paper mainly extends the idea of~\cite{D17-1209}, which regarded encoded vectors of each word as graph nodes and took them with syntactic dependencies as GCN inputs. Following this,~\newcite{N18-2078} obtained better performance by using syntactic and semantic (semantic-role structures) GCNs together, and ~\newcite{P18-1026} improved the representing ability of the encoder through gated GNN with AMR information included. Although they realized that explicit linguistic information could enhance the natural language modeling, their graph node essentially act as a non-recursive quasi-RNNCell in formula, overlooking the internal sequential information between nodes.

In this study, we introduce more flexible strategies for both Recurrent NMT and Transformer, yielding better results than above independent-node graph modelings.

\section{Conclusions and Future Work}

We present a simple yet effective approach, Recurrent Graph Syntax Encoder (RGSE), to inform NMT models with explicit syntactic dependency information. The proposed RGSE is a migratable component on the encoder side which regards RNNCells as graph nodes and injects syntactic dependencies as edges, thereby capturing syntactic information and word order information simultaneously. Our experiments on En-De and En-Cs tasks show that RGSE consistently enhances recurrent NMT~\cite{rnnsearch} and Transformer~\cite{transformer}, achieving the competitive results on par with the SOTA model.

In future work, it will be interesting to apply RGSE to other natural language generation tasks, such as text summarization and conversation.

\bibliography{emnlp-ijcnlp-2019}

\begin{thebibliography}{47}
\expandafter\ifx\csname natexlab\endcsname\relax\def\natexlab#1{#1}\fi

\bibitem[{Aharoni and Goldberg(2017)}]{aharoni-goldberg-2017-towards}
Roee Aharoni and Yoav Goldberg. 2017.
\newblock Towards string-to-tree neural machine translation.
\newblock In \emph{Proceedings of the 55th Annual Meeting of the Association
  for Computational Linguistics (Volume 2: Short Papers)}, pages 132--140,
  Vancouver, Canada. Association for Computational Linguistics.

\bibitem[{Ahmed et~al.(2018)Ahmed, Keskar, and Socher}]{ahmed2018weighted}
Karim Ahmed, Nitish~Shirish Keskar, and Richard Socher. 2018.
\newblock Weighted transformer network for machine translation.

\bibitem[{Anastasopoulos and Chiang(2018)}]{anastasopoulos2018tied}
Antonios Anastasopoulos and David Chiang. 2018.
\newblock Tied multitask learning for neural speech translation.
\newblock In \emph{Proceedings of the 2018 Conference of the North American
  Chapter of the Association for Computational Linguistics: Human Language
  Technologies, Volume 1 (Long Papers)}, pages 82--91.

\bibitem[{Bahdanau et~al.(2015)Bahdanau, Cho, and Bengio}]{rnnsearch}
Dzmitry Bahdanau, KyungHyun Cho, and Yoshua Bengio. 2015.
\newblock Neural machine translation by jointly learning to align and
  translate.
\newblock In \emph{Proceedings of ICLR 2015}.

\bibitem[{Bastings et~al.(2017)Bastings, Titov, Aziz, Marcheggiani, and
  Simaan}]{D17-1209}
Joost Bastings, Ivan Titov, Wilker Aziz, Diego Marcheggiani, and Khalil Simaan.
  2017.
\newblock Graph convolutional encoders for syntax-aware neural machine
  translation.
\newblock In \emph{Proceedings of EMNLP 2017}.

\bibitem[{Battaglia et~al.(2016)Battaglia, Pascanu, Lai, Rezende
  et~al.}]{battaglia2016interaction}
Peter Battaglia, Razvan Pascanu, Matthew Lai, Danilo~Jimenez Rezende, et~al.
  2016.
\newblock Interaction networks for learning about objects, relations and
  physics.
\newblock In \emph{Proceedings of NIPS 2016}, pages 4502--4510.

\bibitem[{Beck et~al.(2018)Beck, Haffari, and Cohn}]{P18-1026}
Daniel Beck, Gholamreza Haffari, and Trevor Cohn. 2018.
\newblock Graph-to-sequence learning using gated graph neural networks.
\newblock In \emph{Proceedings of ACL 2018}.

\bibitem[{Chen et~al.(2017)Chen, Huang, Chiang, and
  Chen}]{chen-EtAl:2017:Long6}
Huadong Chen, Shujian Huang, David Chiang, and Jiajun Chen. 2017.
\newblock Improved neural machine translation with a syntax-aware encoder and
  decoder.
\newblock In \emph{Proceedings of ACL 2017}, pages 1936--1945.

\bibitem[{Domhan(2018)}]{P18-1167}
Tobias Domhan. 2018.
\newblock How much attention do you need? a granular analysis of neural machine
  translation architectures.
\newblock In \emph{Proceedings of ACL 2018}.

\bibitem[{Eriguchi et~al.(2016)Eriguchi, Hashimoto, and
  Tsuruoka}]{eriguchi-hashimoto-tsuruoka:2016:P16-1}
Akiko Eriguchi, Kazuma Hashimoto, and Yoshimasa Tsuruoka. 2016.
\newblock Tree-to-sequence attentional neural machine translation.
\newblock In \emph{Proceedings of ACL 2016}, pages 823--833.

\bibitem[{Gehring et~al.(2017)Gehring, Auli, Grangier, Yarats, and
  Dauphin}]{convs2s}
Jonas Gehring, Michael Auli, David Grangier, Denis Yarats, and Yann~N Dauphin.
  2017.
\newblock Convolutional sequence to sequence learning.
\newblock In \emph{Proceedings of ICML 2017}.

\bibitem[{Hamilton et~al.(2017)Hamilton, Ying, and
  Leskovec}]{hamilton2017inductive}
William~L. Hamilton, Rex Ying, and Jure Leskovec. 2017.
\newblock Inductive representation learning on large graphs.
\newblock In \emph{Proceedings of NIPS 2017}.

\bibitem[{He et~al.(2016)He, Zhang, Ren, and Sun}]{he2016deep}
Kaiming He, Xiangyu Zhang, Shaoqing Ren, and Jian Sun. 2016.
\newblock Deep residual learning for image recognition.
\newblock In \emph{Proceedings of CVPR 2016}, pages 770--778.

\bibitem[{Kalchbrenner and Blunsom(2013)}]{D13-1176}
Nal Kalchbrenner and Phil Blunsom. 2013.
\newblock Recurrent continuous translation models.
\newblock In \emph{Proceedings of EMNLP 2013}, pages 1700--1709.

\bibitem[{Kipf and Welling(2016)}]{kipf2016semi}
Thomas~N Kipf and Max Welling. 2016.
\newblock Semi-supervised classification with graph convolutional networks.
\newblock \emph{arXiv preprint arXiv:1609.02907}.

\bibitem[{Klein et~al.(2017)Klein, Kim, Deng, Senellart, and Rush}]{opennmt}
Guillaume Klein, Yoon Kim, Yuntian Deng, Jean Senellart, and Alexander~M. Rush.
  2017.
\newblock Open{NMT}: Open-source toolkit for neural machine translation.
\newblock In \emph{Proc. ACL}.

\bibitem[{Kuncoro et~al.(2018)Kuncoro, Dyer, Hale, Yogatama, Clark, and
  Blunsom}]{kuncoro2018lstms}
Adhiguna Kuncoro, Chris Dyer, John Hale, Dani Yogatama, Stephen Clark, and Phil
  Blunsom. 2018.
\newblock Lstms can learn syntax-sensitive dependencies well, but modeling
  structure makes them better.
\newblock In \emph{Proceedings of the 56th Annual Meeting of the Association
  for Computational Linguistics (Volume 1: Long Papers)}, pages 1426--1436.

\bibitem[{Li et~al.(2017)Li, Xiong, Tu, Zhu, Zhang, and
  Zhou}]{li-EtAl:2017:Long}
Junhui Li, Deyi Xiong, Zhaopeng Tu, Muhua Zhu, Min Zhang, and Guodong Zhou.
  2017.
\newblock Modeling source syntax for neural machine translation.
\newblock In \emph{Proceedings of ACL 2017}, pages 688--697.

\bibitem[{Linzen et~al.(2016)Linzen, Dupoux, and
  Goldberg}]{linzen2016assessing}
Tal Linzen, Emmanuel Dupoux, and Yoav Goldberg. 2016.
\newblock Assessing the ability of lstms to learn syntax-sensitive
  dependencies.
\newblock \emph{Transactions of the Association for Computational Linguistics},
  4:521--535.

\bibitem[{Luong et~al.(2016)Luong, Le, Sutskever, Vinyals, and
  Kaiser}]{luong2015multitask}
Minh-Thang Luong, Quoc~V. Le, Ilya Sutskever, Oriol Vinyals, and Lukasz Kaiser.
  2016.
\newblock Multi-task sequence to sequence learning.
\newblock In \emph{Proceedings of ICLR 2016}.

\bibitem[{Ma et~al.(2018)Ma, Tamura, Utiyama, Zhao, and Sumita}]{P18-1116}
Chunpeng Ma, Akihiro Tamura, Masao Utiyama, Tiejun Zhao, and Eiichiro Sumita.
  2018.
\newblock Forest-based neural machine translation.
\newblock In \emph{Proceedings of ACL 2018}.

\bibitem[{Marcheggiani et~al.(2018)Marcheggiani, Bastings, and
  Titov}]{N18-2078}
Diego Marcheggiani, Joost Bastings, and Ivan Titov. 2018.
\newblock Exploiting semantics in neural machine translation with graph
  convolutional networks.
\newblock In \emph{Proceedings of NAACL 2018}.

\bibitem[{Marcheggiani and Titov(2017)}]{marcheggiani-titov:2017:EMNLP2017}
Diego Marcheggiani and Ivan Titov. 2017.
\newblock Encoding sentences with graph convolutional networks for semantic
  role labeling.
\newblock In \emph{Proceedings of EMNLP 2017}, pages 1506--1515.

\bibitem[{Niehues and Cho(2017)}]{Niehues2017ExploitingLR}
Jan Niehues and Eunah Cho. 2017.
\newblock Exploiting linguistic resources for neural machine translation using
  multi-task learning.
\newblock In \emph{Proceedings of the WMT 2017}.

\bibitem[{Papineni et~al.(2002)Papineni, Roukos, Ward, and
  Zhu}]{Papineni:2002:BMA:1073083.1073135}
Kishore Papineni, Salim Roukos, Todd Ward, and Wei-Jing Zhu. 2002.
\newblock Bleu: A method for automatic evaluation of machine translation.
\newblock In \emph{Proceedings of ACL 2002}, pages 311--318.

\bibitem[{Raganato and Tiedemann(2018)}]{raganato2018analysis}
Alessandro Raganato and J{\"o}rg Tiedemann. 2018.
\newblock An analysis of encoder representations in transformer-based machine
  translation.
\newblock In \emph{Proceedings of the 2018 EMNLP Workshop BlackboxNLP:
  Analyzing and Interpreting Neural Networks for NLP}, pages 287--297.

\bibitem[{Scarselli et~al.(2009)Scarselli, Gori, Tsoi, Hagenbuchner, and
  Monfardini}]{scarselli2009graph}
Franco Scarselli, Marco Gori, Ah~Chung Tsoi, Markus Hagenbuchner, and Gabriele
  Monfardini. 2009.
\newblock The graph neural network model.
\newblock \emph{IEEE Transactions on Neural Networks}, 20(1):61--80.

\bibitem[{Sennrich and Haddow(2016)}]{sennrich-haddow:2016:WMT}
Rico Sennrich and Barry Haddow. 2016.
\newblock Linguistic input features improve neural machine translation.
\newblock In \emph{Proceedings of the WMT 2016}, pages 83--91.

\bibitem[{Sennrich et~al.(2016)Sennrich, Haddow, and
  Birch}]{sennrich2016neural}
Rico Sennrich, Barry Haddow, and Alexandra Birch. 2016.
\newblock Neural machine translation of rare words with subword units.
\newblock In \emph{Proceedings of the 54th Annual Meeting of the Association
  for Computational Linguistics (Volume 1: Long Papers)}, volume~1, pages
  1715--1725.

\bibitem[{Shaw et~al.(2018)Shaw, Uszkoreit, and Vaswani}]{shaw2018self}
Peter Shaw, Jakob Uszkoreit, and Ashish Vaswani. 2018.
\newblock Self-attention with relative position representations.
\newblock In \emph{Proceedings of the 2018 Conference of the North American
  Chapter of the Association for Computational Linguistics: Human Language
  Technologies, Volume 2 (Short Papers)}, pages 464--468.

\bibitem[{Shi et~al.(2016)Shi, Padhi, and Knight}]{shi2016does}
Xing Shi, Inkit Padhi, and Kevin Knight. 2016.
\newblock Does string-based neural mt learn source syntax?
\newblock In \emph{Proceedings of the 2016 Conference on Empirical Methods in
  Natural Language Processing}, pages 1526--1534.

\bibitem[{Song et~al.(2019)Song, Gildea, Zhang, Wang, and
  Su}]{song2019semantic}
Linfeng Song, Daniel Gildea, Yue Zhang, Zhiguo Wang, and Jinsong Su. 2019.
\newblock Semantic neural machine translation using amr.
\newblock \emph{arXiv preprint arXiv:1902.07282}.

\bibitem[{Song et~al.(2018{\natexlab{a}})Song, Zhang, Wang, and
  Gildea}]{song2018graph}
Linfeng Song, Yue Zhang, Zhiguo Wang, and Daniel Gildea. 2018{\natexlab{a}}.
\newblock A graph-to-sequence model for amr-to-text generation.
\newblock In \emph{Proceedings of ACL 2018}.

\bibitem[{Song et~al.(2018{\natexlab{b}})Song, Zhang, Wang, and
  Gildea}]{song2018n}
Linfeng Song, Yue Zhang, Zhiguo Wang, and Daniel Gildea. 2018{\natexlab{b}}.
\newblock N-ary relation extraction using graph state lstm.
\newblock In \emph{Proceedings of EMNLP 2018}.

\bibitem[{Stahlberg et~al.(2016)Stahlberg, Hasler, Waite, and
  Byrne}]{stahlberg-etal-2016-syntactically}
Felix Stahlberg, Eva Hasler, Aurelien Waite, and Bill Byrne. 2016.
\newblock Syntactically guided neural machine translation.
\newblock In \emph{Proceedings of the 54th Annual Meeting of the Association
  for Computational Linguistics (Volume 2: Short Papers)}, pages 299--305,
  Berlin, Germany. Association for Computational Linguistics.

\bibitem[{Strubell et~al.(2018)Strubell, Verga, Andor, Weiss, and
  McCallum}]{strubell2018linguistically}
Emma Strubell, Patrick Verga, Daniel Andor, David Weiss, and Andrew McCallum.
  2018.
\newblock Linguistically-informed self-attention for semantic role labeling.
\newblock \emph{arXiv preprint arXiv:1804.08199}.

\bibitem[{Sutskever et~al.(2014)Sutskever, Vinyals, and Le}]{seq2seq}
Ilya Sutskever, Oriol Vinyals, and Quoc~V Le. 2014.
\newblock Sequence to sequence learning with neural networks.
\newblock In \emph{Proceedings of NIPS 2014}, pages 3104--3112.

\bibitem[{Tai et~al.(2015)Tai, Socher, and Manning}]{Tai2015ImprovedSR}
Kai~Sheng Tai, Richard Socher, and Christopher~D. Manning. 2015.
\newblock Improved semantic representations from tree-structured long
  short-term memory networks.
\newblock In \emph{Proceedings of ACL 2015}.

\bibitem[{Vaswani et~al.(2017)Vaswani, Shazeer, Parmar, Uszkoreit, Jones,
  Gomez, Kaiser, and Polosukhin}]{transformer}
Ashish Vaswani, Noam Shazeer, Niki Parmar, Jakob Uszkoreit, Llion Jones,
  Aidan~N Gomez, Lukasz Kaiser, and Illia Polosukhin. 2017.
\newblock Attention is all you need.
\newblock In \emph{Proceedings of NIPS 2017}.

\bibitem[{Wu et~al.(2019)Wu, Fan, Baevski, Dauphin, and Auli}]{wu2019pay}
Felix Wu, Angela Fan, Alexei Baevski, Yann~N Dauphin, and Michael Auli. 2019.
\newblock Pay less attention with lightweight and dynamic convolutions.
\newblock \emph{arXiv preprint arXiv:1901.10430}.

\bibitem[{Wu et~al.(2018)Wu, Zhang, Zhang, Yang, Li, and
  Zhou}]{Wu:2018:DNM:3281228.3281242}
Shuangzhi Wu, Dongdong Zhang, Zhirui Zhang, Nan Yang, Mu~Li, and Ming Zhou.
  2018.
\newblock Dependency-to-dependency neural machine translation.
\newblock \emph{IEEE/ACM Trans. Audio, Speech and Lang. Proc.},
  26(11):2132--2141.

\bibitem[{Wu et~al.(2017)Wu, Zhou, and Zhang}]{wu2017improved}
Shuangzhi Wu, Ming Zhou, and Dongdong Zhang. 2017.
\newblock Improved neural machine translation with source syntax.
\newblock In \emph{IJCAI}, pages 4179--4185.

\bibitem[{Wu et~al.(2016)Wu, Schuster, Chen, Le, Norouzi, Macherey, Krikun,
  Cao, Gao, Macherey, Klingner, Shah, Johnson et~al.}]{gnmt}
Yonghui Wu, Mike Schuster, Zhifeng Chen, Quoc~V. Le, Mohammad Norouzi, Wolfgang
  Macherey, Maxim Krikun, Yuan Cao, Qin Gao, Klaus Macherey, Jeff Klingner,
  Apurva Shah, Melvin Johnson, et~al. 2016.
\newblock Google's neural machine translation system: Bridging the gap between
  human and machine translation.
\newblock In \emph{Proceedings of NIPS 2016}.

\bibitem[{Yang et~al.(2018)Yang, Tu, Wong, Meng, Chao, and
  Zhang}]{yang2018modeling}
Baosong Yang, Zhaopeng Tu, Derek~F Wong, Fandong Meng, Lidia~S Chao, and Tong
  Zhang. 2018.
\newblock Modeling localness for self-attention networks.
\newblock In \emph{Proceedings of the 2018 Conference on Empirical Methods in
  Natural Language Processing}, pages 4449--4458.

\bibitem[{Zaremoodi and Haffari(2018)}]{C18-1120}
Poorya Zaremoodi and Gholamreza Haffari. 2018.
\newblock Incorporating syntactic uncertainty in neural machine translation
  with a forest-to-sequence model.
\newblock In \emph{Proceedings of COLING 2018}, pages 1421--1429.

\bibitem[{Zhang et~al.(2019)Zhang, Li, Fu, and Zhang}]{zhang2019syntax}
Meishan Zhang, Zhenghua Li, Guohong Fu, and Min Zhang. 2019.
\newblock Syntax-enhanced neural machine translation with syntax-aware word
  representations.
\newblock \emph{arXiv preprint arXiv:1905.02878}.

\bibitem[{Zhang et~al.(2018)Zhang, Liu, and Song}]{zhang2018sentence}
Yue Zhang, Qi~Liu, and Linfeng Song. 2018.
\newblock Sentence-state lstm for text representation.
\newblock In \emph{Proceedings of the 56th Annual Meeting of the Association
  for Computational Linguistics (Volume 1: Long Papers)}, pages 317--327.

\end{thebibliography}
\bibliographystyle{acl_natbib}

\appendix

\end{document}